\documentclass[11pt]{article}

\newenvironment{descit}[1]{\begin{quote}\noindent\textit{#1}}{\end{quote}}

\ifx\undefined\psfig\else \fi

\edef\psfigRestoreAt{\catcode`@=\number\catcode`@\relax}
\catcode`\@=11\relax
\newwrite\@unused
\def\typeout#1{{\let\protect\string\immediate\write\@unused{#1}}}
\def\figurepath{./}

\def\@nnil{\@nil}
\def\@empty{}
\def\@psdonoop#1\@@#2#3{}
\def\@psdo#1:=#2\do#3{\edef\@psdotmp{#2}\ifx\@psdotmp\@empty \else
    \expandafter\@psdoloop#2,\@nil,\@nil\@@#1{#3}\fi}
\def\@psdoloop#1,#2,#3\@@#4#5{\def#4{#1}\ifx #4\@nnil \else
       #5\def#4{#2}\ifx #4\@nnil \else#5\@ipsdoloop #3\@@#4{#5}\fi\fi}
\def\@ipsdoloop#1,#2\@@#3#4{\def#3{#1}\ifx #3\@nnil 
       \let\@nextwhile=\@psdonoop \else
      #4\relax\let\@nextwhile=\@ipsdoloop\fi\@nextwhile#2\@@#3{#4}}
\def\@tpsdo#1:=#2\do#3{\xdef\@psdotmp{#2}\ifx\@psdotmp\@empty \else
    \@tpsdoloop#2\@nil\@nil\@@#1{#3}\fi}
\def\@tpsdoloop#1#2\@@#3#4{\def#3{#1}\ifx #3\@nnil 
       \let\@nextwhile=\@psdonoop \else
      #4\relax\let\@nextwhile=\@tpsdoloop\fi\@nextwhile#2\@@#3{#4}}
\newread\ps@stream
\newif\ifnot@eof       % continue looking for the bounding box?
\newif\if@noisy        % report what you're making?
\newif\if@atend        % %%BoundingBox: has (at end) specification
\newif\if@psfile       % does this look like a PostScript file?
{\catcode`\%=12\global\gdef\epsf@start{%!}}
\def\epsf@PS{PS}
\def\epsf@getbb#1{%
\openin\ps@stream=#1
\ifeof\ps@stream\typeout{Error, File #1 not found}\else
   {\not@eoftrue \chardef\other=12
    \def\do##1{\catcode`##1=\other}\dospecials \catcode`\ =10
    \loop
       \if@psfile
	  \read\ps@stream to \epsf@fileline
       \else{
	  \obeyspaces
          \read\ps@stream to \epsf@tmp\global\let\epsf@fileline\epsf@tmp}
       \fi
       \ifeof\ps@stream\not@eoffalse\else
       \if@psfile\else
       \expandafter\epsf@test\epsf@fileline:. \\%
       \fi
          \expandafter\epsf@aux\epsf@fileline:. \\%
       \fi
   \ifnot@eof\repeat
   }\closein\ps@stream\fi}%
\long\def\epsf@test#1#2#3:#4\\{\def\epsf@testit{#1#2}
			\ifx\epsf@testit\epsf@start\else
\typeout{Warning! File does not start with `\epsf@start'.  It may not be a PostScript file.}
			\fi
			\@psfiletrue} % don't test after 1st line
{\catcode`\%=12\global\let\epsf@percent=%\global\def\epsf@bblit{%BoundingBox}}
\long\def\epsf@aux#1#2:#3\\{\ifx#1\epsf@percent
   \def\epsf@testit{#2}\ifx\epsf@testit\epsf@bblit
	\@atendfalse
        \epsf@atend #3 . \\%
	\if@atend	
	   \if@verbose{
		\typeout{psfig: found `(atend)'; continuing search}
	   }\fi
        \else
        \epsf@grab #3 . . . \\%
        \not@eoffalse
        \global\no@bbfalse
        \fi
   \fi\fi}%
\def\epsf@grab #1 #2 #3 #4 #5\\{%
   \global\def\epsf@llx{#1}\ifx\epsf@llx\empty
      \epsf@grab #2 #3 #4 #5 .\\\else
   \global\def\epsf@lly{#2}%
   \global\def\epsf@urx{#3}\global\def\epsf@ury{#4}\fi}%
\def\epsf@atendlit{(atend)} 
\def\epsf@atend #1 #2 #3\\{%
   \def\epsf@tmp{#1}\ifx\epsf@tmp\empty
      \epsf@atend #2 #3 .\\\else
   \ifx\epsf@tmp\epsf@atendlit\@atendtrue\fi\fi}

\chardef\letter = 11
\chardef\other = 12

\newif \ifdebug %%% turn me on to see TeX hard at work ...
\newif\ifc@mpute %%% don't need to compute some values
\c@mputetrue % but assume that we do

\let\then = \relax
\def\r@dian{pt }
\let\r@dians = \r@dian
\let\dimensionless@nit = \r@dian
\let\dimensionless@nits = \dimensionless@nit
\def\internal@nit{sp }
\let\internal@nits = \internal@nit
\newif\ifstillc@nverging
\def \Mess@ge #1{\ifdebug \then \message {#1} \fi}

{ %%% Things that need abnormal catcodes %%%
	\catcode `\@ = \letter
	\gdef \nodimen {\expandafter \n@dimen \the \dimen}
	\gdef \term #1 #2 #3%
	       {\edef \t@ {\the #1}%%% freeze parameter 1 (count, by value)
		\edef \t@@ {\expandafter \n@dimen \the #2\r@dian}%
		\t@rm {\t@} {\t@@} {#3}%
	       }
	\gdef \t@rm #1 #2 #3%
	       {{%
		\count 0 = 0
		\dimen 0 = 1 \dimensionless@nit
		\dimen 2 = #2\relax
		\Mess@ge {Calculating term #1 of \nodimen 2}%
		\loop
		\ifnum	\count 0 < #1
		\then	\advance \count 0 by 1
			\Mess@ge {Iteration \the \count 0 \space}%
			\Multiply \dimen 0 by {\dimen 2}%
			\Mess@ge {After multiplication, term = \nodimen 0}%
			\Divide \dimen 0 by {\count 0}%
			\Mess@ge {After division, term = \nodimen 0}%
		\repeat
		\Mess@ge {Final value for term #1 of 
				\nodimen 2 \space is \nodimen 0}%
		\xdef \Term {#3 = \nodimen 0 \r@dians}%
		\aftergroup \Term
	       }}
	\catcode `\p = \other
	\catcode `\t = \other
	\gdef \n@dimen #1pt{#1} %%% throw away the ``pt''
}

\def \Divide #1by #2{\divide #1 by #2} %%% just a synonym

\def \Multiply #1by #2%%% allows division of a dimen by a dimen
       {{%%% should really freeze parameter 2 (dimen, passed by value)
	\count 0 = #1\relax
	\count 2 = #2\relax
	\count 4 = 65536
	\Mess@ge {Before scaling, count 0 = \the \count 0 \space and
			count 2 = \the \count 2}%
	\ifnum	\count 0 > 32767 %%% do our best to avoid overflow
	\then	\divide \count 0 by 4
		\divide \count 4 by 4
	\else	\ifnum	\count 0 < -32767
		\then	\divide \count 0 by 4
			\divide \count 4 by 4
		\else
		\fi
	\fi
	\ifnum	\count 2 > 32767 %%% while retaining reasonable accuracy
	\then	\divide \count 2 by 4
		\divide \count 4 by 4
	\else	\ifnum	\count 2 < -32767
		\then	\divide \count 2 by 4
			\divide \count 4 by 4
		\else
		\fi
	\fi
	\multiply \count 0 by \count 2
	\divide \count 0 by \count 4
	\xdef \product {#1 = \the \count 0 \internal@nits}%
	\aftergroup \product
       }}

\def\r@duce{\ifdim\dimen0 > 90\r@dian \then   % sin(x) = sin(180-x)
		\multiply\dimen0 by -1
		\advance\dimen0 by 180\r@dian
		\r@duce
	    \else \ifdim\dimen0 < -90\r@dian \then  % sin(x) = sin(360+x)
		\advance\dimen0 by 360\r@dian
		\r@duce
		\fi
	    \fi}

\def\Sine#1%
       {{%
	\dimen 0 = #1 \r@dian
	\r@duce
	\ifdim\dimen0 = -90\r@dian \then
	   \dimen4 = -1\r@dian
	   \c@mputefalse
	\fi
	\ifdim\dimen0 = 90\r@dian \then
	   \dimen4 = 1\r@dian
	   \c@mputefalse
	\fi
	\ifdim\dimen0 = 0\r@dian \then
	   \dimen4 = 0\r@dian
	   \c@mputefalse
	\fi
	\ifc@mpute \then
		\divide\dimen0 by 180
		\dimen0=3.141592654\dimen0
		\dimen 2 = 3.1415926535897963\r@dian %%% a well-known constant
		\divide\dimen 2 by 2 %%% we only deal with -pi/2 : pi/2
		\Mess@ge {Sin: calculating Sin of \nodimen 0}%
		\count 0 = 1 %%% see power-series expansion for sine
		\dimen 2 = 1 \r@dian %%% ditto
		\dimen 4 = 0 \r@dian %%% ditto
		\loop
			\ifnum	\dimen 2 = 0 %%% then we've done
			\then	\stillc@nvergingfalse 
			\else	\stillc@nvergingtrue
			\fi
			\ifstillc@nverging %%% then calculate next term
			\then	\term {\count 0} {\dimen 0} {\dimen 2}%
				\advance \count 0 by 2
				\count 2 = \count 0
				\divide \count 2 by 2
				\ifodd	\count 2 %%% signs alternate
				\then	\advance \dimen 4 by \dimen 2
				\else	\advance \dimen 4 by -\dimen 2
				\fi
		\repeat
	\fi		
			\xdef \sine {\nodimen 4}%
       }}

\def\Cosine#1{\ifx\sine\UnDefined\edef\Savesine{\relax}\else
		             \edef\Savesine{\sine}\fi
	{\dimen0=#1\r@dian\multiply\dimen0 by -1
	 \advance\dimen0 by 90\r@dian
	 \Sine{\nodimen 0}
	 \xdef\cosine{\sine}
	 \xdef\sine{\Savesine}}}	      
\def\psdraft{
	\def\@psdraft{0}
}
\def\psfull{
	\def\@psdraft{100}
}

\psfull

\newif\if@draftbox
\def\psnodraftbox{
	\@draftboxfalse
}
\@draftboxtrue

\newif\if@prologfile
\newif\if@postlogfile
\def\pssilent{
	\@noisyfalse
}
\def\psnoisy{
	\@noisytrue
}
\psnoisy
\newif\if@bbllx
\newif\if@bblly
\newif\if@bburx
\newif\if@bbury
\newif\if@height
\newif\if@width
\newif\if@rheight
\newif\if@rwidth
\newif\if@angle
\newif\if@clip
\newif\if@verbose
\newif\if@scale
\def\@p@@sclip#1{\@cliptrue}

\def\@p@@sfile#1{\def\@p@sfile{null}%
	        \openin1=#1
		\ifeof1\closein1%
		       \openin1=\figurepath#1
			\ifeof1\typeout{Error, File #1 not found}
			   \if@bbllx\if@bblly\if@bburx\if@bbury% added 6/91 Rob Russell
			      \def\@p@sfile{#1}%
			   \fi\fi\fi\fi
			\else\closein1
			    \edef\@p@sfile{\figurepath#1}%
                        \fi%
		 \else\closein1%
		       \def\@p@sfile{#1}%
		 \fi}
\def\@p@@sfigure#1{\def\@p@sfile{null}%
	        \openin1=#1
		\ifeof1\closein1%
		       \openin1=\figurepath#1
			\ifeof1\typeout{Error, File #1 not found}
			   \if@bbllx\if@bblly\if@bburx\if@bbury% added 6/91 Rob Russell
			      \def\@p@sfile{#1}%
			   \fi\fi\fi\fi
			\else\closein1
			    \def\@p@sfile{\figurepath#1}%
                        \fi%
		 \else\closein1%
		       \def\@p@sfile{#1}%
		 \fi}

\def\@p@@sbbllx#1{
		\@bbllxtrue
		\dimen100=#1
		\edef\@p@sbbllx{\number\dimen100}
}
\def\@p@@sbblly#1{
		\@bbllytrue
		\dimen100=#1
		\edef\@p@sbblly{\number\dimen100}
}
\def\@p@@sbburx#1{
		\@bburxtrue
		\dimen100=#1
		\edef\@p@sbburx{\number\dimen100}
}
\def\@p@@sbbury#1{
		\@bburytrue
		\dimen100=#1
		\edef\@p@sbbury{\number\dimen100}
}
\def\@p@@sheight#1{
		\@heighttrue
		\dimen100=#1
   		\edef\@p@sheight{\number\dimen100}
}
\def\@p@@swidth#1{
		\@widthtrue
		\dimen100=#1
		\edef\@p@swidth{\number\dimen100}
}
\def\@p@@srheight#1{
		\@rheighttrue
		\dimen100=#1
		\edef\@p@srheight{\number\dimen100}
}
\def\@p@@srwidth#1{
		\@rwidthtrue
		\dimen100=#1
		\edef\@p@srwidth{\number\dimen100}
}
\def\@p@@sangle#1{
		\@angletrue
		\edef\@p@sangle{#1} %\number\dimen100}
}
\def\@p@@ssilent#1{ 
		\@verbosefalse
}
\def\@p@@sscale#1{
		\def\@p@scale{#1}
		\@scaletrue
}
\def\@p@@sprolog#1{\@prologfiletrue\def\@prologfileval{#1}}
\def\@p@@spostlog#1{\@postlogfiletrue\def\@postlogfileval{#1}}
\def\@cs@name#1{\csname #1\endcsname}
\def\@setparms#1=#2,{\@cs@name{@p@@s#1}{#2}}
\def\ps@init@parms{
		\@bbllxfalse \@bbllyfalse
		\@bburxfalse \@bburyfalse
		\@heightfalse \@widthfalse
		\@rheightfalse \@rwidthfalse
		\@scalefalse
		\def\@p@sbbllx{}\def\@p@sbblly{}
		\def\@p@sbburx{}\def\@p@sbbury{}
		\def\@p@sheight{}\def\@p@swidth{}
		\def\@p@srheight{}\def\@p@srwidth{}
		\def\@p@sangle{0}
		\def\@p@sfile{}
		\def\@p@scost{10}
		\def\@sc{}
		\@prologfilefalse
		\@postlogfilefalse
		\@clipfalse
		\if@noisy
			\@verbosetrue
		\else
			\@verbosefalse
		\fi
}
\def\parse@ps@parms#1{
	 	\@psdo\@psfiga:=#1\do
		   {\expandafter\@setparms\@psfiga,}}
\newif\ifno@bb
\def\bb@missing{
	\if@verbose{
		\typeout{psfig: searching \@p@sfile \space  for bounding box}
	}\fi
	\no@bbtrue
	\epsf@getbb{\@p@sfile}
        \ifno@bb \else \bb@cull\epsf@llx\epsf@lly\epsf@urx\epsf@ury\fi
}	
\def\bb@cull#1#2#3#4{
	\dimen100=#1 bp\edef\@p@sbbllx{\number\dimen100}
	\dimen100=#2 bp\edef\@p@sbblly{\number\dimen100}
	\dimen100=#3 bp\edef\@p@sbburx{\number\dimen100}
	\dimen100=#4 bp\edef\@p@sbbury{\number\dimen100}
	\no@bbfalse
}

\newdimen\p@intvaluex
\newdimen\p@intvaluey
\newdimen\@ffsetvalue
\newdimen\x@ffsetvalue
\newdimen\y@ffsetvalue

\def\compute@offset#1#2{{\dimen0=#1 sp\dimen1=#2 sp
			\advance\dimen1 by -\dimen0
			\dimen1=\sine\dimen1
			\dimen0=\cosine\dimen1
			\ifdim\dimen0<0sp \dimen1=0sp \fi
			\global\@ffsetvalue=\dimen1}}

\def\rotate@#1#2{{\dimen0=#1 sp\dimen1=#2 sp
		  \global\p@intvaluex=\cosine\dimen0
		  \dimen3=\sine\dimen1
		  \global\advance\p@intvaluex by -\dimen3
		  \global\p@intvaluey=\sine\dimen0
		  \dimen3=\cosine\dimen1
		  \global\advance\p@intvaluey by \dimen3
		  }}
\def\compute@bb{
		\no@bbfalse
		\if@bbllx \else \no@bbtrue \fi
		\if@bblly \else \no@bbtrue \fi
		\if@bburx \else \no@bbtrue \fi
		\if@bbury \else \no@bbtrue \fi
		\ifno@bb \bb@missing \fi
		\ifno@bb \typeout{FATAL ERROR: no bb supplied or found}
			\no-bb-error
		\fi
		\if@angle 
			\Sine{\@p@sangle}\Cosine{\@p@sangle}
			\compute@offset{\@p@sbblly}{\@p@sbbury}
			\x@ffsetvalue=\@ffsetvalue
			\compute@offset{\@p@sbburx}{\@p@sbbllx}
			\y@ffsetvalue=\@ffsetvalue

			\rotate@{\@p@sbbllx}{\@p@sbblly}
			\advance\p@intvaluex by -\x@ffsetvalue
			\advance\p@intvaluey by -\y@ffsetvalue
			\edef\@p@sbbllx{\number\p@intvaluex}
			\edef\@p@sbblly{\number\p@intvaluey}

			\rotate@{\@p@sbburx}{\@p@sbbury}
			\advance\p@intvaluex by \x@ffsetvalue
			\advance\p@intvaluey by \y@ffsetvalue
			\edef\@p@sbburx{\number\p@intvaluex}
			\edef\@p@sbbury{\number\p@intvaluey}
			{
			 \count0=\@p@sbbllx \count1=\@p@sbblly
		 	 \count2=\@p@sbburx \count3=\@p@sbbury
			 \dimen0=\@p@sbbllx sp\dimen1=\@p@sbblly sp
		 	 \dimen2=\@p@sbburx sp\dimen3=\@p@sbbury sp
			 \dimen203=\dimen2 \advance\dimen203 by -\dimen0
			 \dimen204=\dimen3 \advance\dimen204 by -\dimen1
			 \ifdim\dimen203<0sp 
			      \count203=\count2 \count2=\count0 
			      \count0=\count203 
			      \global\edef\@p@sbbllx{\number\count0}
			      \global\edef\@p@sbburx{\number\count2}
			 \fi
			 \ifdim\dimen204<0sp 
			       \count204=\count3
			       \count3=\count1
			       \count1=\count204
			       \global\edef\@p@sbblly{\number\count1}
			       \global\edef\@p@sbbury{\number\count3}
			 \fi
			}
		\fi
		\count203=\@p@sbburx
		\count204=\@p@sbbury
		\advance\count203 by -\@p@sbbllx
		\advance\count204 by -\@p@sbblly
		\edef\@bbw{\number\count203}
		\edef\@bbh{\number\count204}
}
\def\in@hundreds#1#2#3{\count240=#2 \count241=#3
		     \count100=\count240	% 100 is first digit #2/#3
		     \divide\count100 by \count241
		     \count101=\count100
		     \multiply\count101 by \count241
		     \advance\count240 by -\count101
		     \multiply\count240 by 10
		     \count101=\count240	%101 is second digit of #2/#3
		     \divide\count101 by \count241
		     \count102=\count101
		     \multiply\count102 by \count241
		     \advance\count240 by -\count102
		     \multiply\count240 by 10
		     \count102=\count240	% 102 is the third digit
		     \divide\count102 by \count241
		     \count200=#1\count205=0
		     \count201=\count200
			\multiply\count201 by \count100
		 	\advance\count205 by \count201
		     \count201=\count200
			\divide\count201 by 10
			\multiply\count201 by \count101
			\advance\count205 by \count201
		     \count201=\count200
			\divide\count201 by 100
			\multiply\count201 by \count102
			\advance\count205 by \count201
		     \edef\@result{\number\count205}
}
\def\@ScaleInHundreds#1{
		\in@hundreds{#1}{\@p@scale}{100}
		\edef#1{\@result}
}
\def\compute@wfromh{
		\in@hundreds{\@p@sheight}{\@bbw}{\@bbh}
		\edef\@p@swidth{\@result}
}
\def\compute@hfromw{
		\in@hundreds{\@p@swidth}{\@bbh}{\@bbw}
		\edef\@p@sheight{\@result}
}
\def\compute@handw{
		\if@height 
			\if@width
			\else
				\compute@wfromh
			\fi
		\else 
			\if@width
				\compute@hfromw
			\else
				\edef\@p@sheight{\@bbh}
				\edef\@p@swidth{\@bbw}
			\fi
		\fi
}
\def\compute@resv{
		\if@rheight \else \edef\@p@srheight{\@p@sheight} \fi
		\if@rwidth \else \edef\@p@srwidth{\@p@swidth} \fi
}
\def\compute@sizes{
	\compute@bb
	\compute@handw
	\compute@resv
}
\def\psfig#1{\vbox {
	\ps@init@parms
	\parse@ps@parms{#1}
	\compute@sizes
	\if@scale
                \if@verbose
                        \typeout{psfig: scaling by \@p@scale}
                \fi
                \@ScaleInHundreds{\@p@swidth}
                \@ScaleInHundreds{\@p@sheight}
                \@ScaleInHundreds{\@p@srwidth}
                \@ScaleInHundreds{\@p@srheight}
        \fi
	\ifnum\@p@scost<\@psdraft{
		\if@verbose{
			\typeout{psfig: including \@p@sfile \space }
		}\fi
		\special{ps::[begin] 	\@p@swidth \space \@p@sheight \space
				\@p@sbbllx \space \@p@sbblly \space
				\@p@sbburx \space \@p@sbbury \space
				startTexFig \space }
		\if@angle
			\special {ps:: \@p@sangle \space rotate \space} 
		\fi
		\if@clip{
			\if@verbose{
				\typeout{(clip)}
			}\fi
			\special{ps:: doclip \space }
		}\fi
		\if@prologfile
		    \special{ps: plotfile \@prologfileval \space } \fi
		\special{ps: plotfile \@p@sfile \space }
		\if@postlogfile
		    \special{ps: plotfile \@postlogfileval \space } \fi
		\special{ps::[end] endTexFig \space }
		\vbox to \@p@srheight true sp{
			\hbox to \@p@srwidth true sp{
				\hss
			}
		\vss
		}
	}\else{
		\if@draftbox{		
			\hbox{\fbox{\vbox to \@p@srheight true sp{
			\vss
			\hbox to \@p@srwidth true sp{ \hss \@p@sfile \hss }
			\vss
			}}}
		}\else{
			\vbox to \@p@srheight true sp{
			\vss
			\hbox to \@p@srwidth true sp{\hss}
			\vss
			}
		}\fi

	}\fi
}}
\def\psglobal{\typeout{psfig: PSGLOBAL is OBSOLETE; use psprint -m instead}}
\psfigRestoreAt

\newif\ifpdf
\ifx\pdfoutput\undefined
  \pdffalse
\else
  \pdfoutput=1
  \pdftrue
\fi

\ifpdf
  \usepackage[pdftex]{graphicx}
  \usepackage[pdftex]{color}
  \DeclareGraphicsExtensions{.pdf,.png,.jpg}
\else
  \usepackage[dvips]{graphicx}
  \usepackage[dvips]{color}
  \DeclareGraphicsExtensions{.eps,.epsi,.ps}
\fi

\usepackage{times}

\def\midv{\mathop{\,|\,}}

\long\def\cbk#1{{\color{red}[CBK: #1]}}
\newlength\colwidth \setlength\colwidth{3.25in}

\title{The Traits of the Personable}
\author{Naren Ramakrishnan\\
Department of Computer Science\\
Virginia Tech, VA 24061, USA\\
Email: naren@cs.vt.edu}
\date{}
\begin{document}

\maketitle
\begin{abstract}
\noindent
Information personalization is fertile ground for application
of AI techniques. In this article I relate personalization to the ability 
to capture partial information in an information-seeking interaction.
The specific focus is on personalizing interactions at web sites.
Using ideas from partial evaluation and explanation-based generalization,
I present a modeling methodology for reasoning about personalization.
This approach helps identify seven tiers of `personable traits' in web sites.
\end{abstract}
%\pagestyle{empty}

%\tableofcontents
%\newpage
\section{Introduction}
Personalization refers to the automatic adjustment of information content,
structure, and presentation tailored to an individual user.
It is increasingly employed by commercial sites to help retain 
customers and reduce information overload. 
For instance, 
Riedl~\cite{riedl-editorial} estimates that there are at least 
23 different types of personalization at Amazon's e-commerce 
site!

What does it mean for a web site to be {\it personable}? 
There are definitely multiple interpretations 
of `personalization'~\cite{cacm-personal} and 
correspondingly many ways of answering this question.
Let us start with the working assumption that a site is personable if it allows
a user's information seeking goals to be met {\it effectively}. Obviously, 
different users have different goals and expectations; thus,
a site that is personable 
for one information-seeking activity might be unpersonable for another.
The goal of this paper is to formalize these intuitive notions and describe
a modeling methodology for reasoning about personalization. 

\subsection{Example}
I will begin with a simple example and later expand the scope of
the discussion to cover more complex cases. A user's interaction with 
a web site can be thought of as a dialog between the user and the underlying
information system, using the communication facilities 
afforded by the web site. Thus, when the user clicks on
a hyperlink or submits data in a form, information is implicitly 
communicated from the user to the system. In response, the system presents
information back to the user (including opportunities for further
user input). Many such dialogs happen in a browsing context.
Consider two users visiting the web site of a camera
retailer:
\vspace{-0.3in}
\begin{descit}{}
\begin{description}
\item [User 1:] I am interested in Nikon models.
\vspace{-0.05in}
\item [User 2:] I am looking for an SLR camera.
\end{description}
\end{descit}

\begin{figure}
\centering
\begin{tabular}{cc}
& \mbox{\psfig{figure=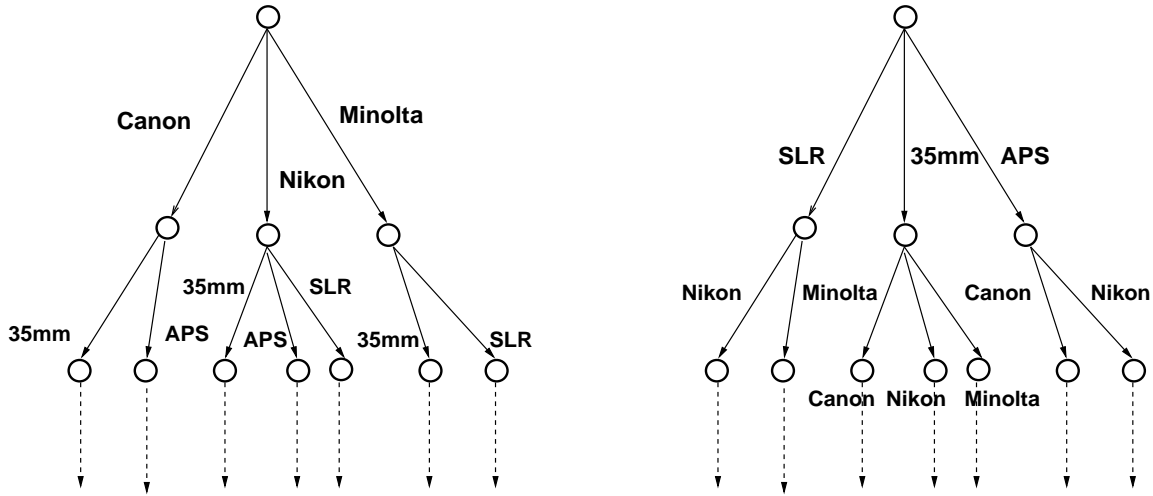,width=6in}}
\end{tabular}
\caption{Two organizations of a camera catalog: by 
maker-type (left) and by type-maker (right). Only two levels are shown
for ease of illustration. The nodes are web pages, the edges denote hyperlinks,
and labels on edges represent the text anchoring the hyperlinks.}
\label{cameras}
\end{figure}

\noindent
User 1 thinks of cameras primarily in terms of their manufacturer and
hence a web site organization such as shown in Fig.~\ref{cameras} (left)
might be appropriate for this user. User 2 thinks of cameras in terms of
lens equipment and Fig.~\ref{cameras} (right) might be preferable.
Both site designs involve a hierarchical browsing
paradigm, with levels corresponding to camera attributes.
Only the top two levels are shown in Fig.~\ref{cameras}; 
it is assumed that further levels impose additional classifications
and distinctions among camera products.

We say that the site in Fig.~\ref{cameras} (left) is personable for 
User 1's activity since it allows the specification of choice of model
(Nikon) at the outset. Conversely, it is {\it unpersonable} for 
User 2's activity
since there is no way for this user to specify her interest in an SLR
camera at the outset. The only option is to follow all possible 
model choices, and determine the ones that support SLR features. This
might involve pursuing paths that are dead-ends (for instance, 
no SLR models are available under Canon). 
In contrast, the organization in Fig.~\ref{cameras} (right) is personable for
User 2's activity since it will eliminate Canon as a potential option
once the selection of SLR is made at the outset. By a symmetric argument
this site is unpersonable for User 1's activity.

Is it possible to have a web site that is personable for both users? The
easy solution is to support two different browsing interfaces --- 
browse-by-maker and browse-by-type. The responsibility is then on the user to
employ the right interface for his activity. 

There are two problems with such a design. The first is the
explosion in the scenario possibilities. If cameras are distinguished by
$n$ independent attributes, then we have $n$! possible organizations. Some web
sites actually take such an exhaustive approach and support all possible
ways of interacting with an information system (e.g., epicurious.com,
a web site for organizing recipes, described in~\cite{hearst-setting}).
The more fundamental problem with such a design is that it over-specifies
the personalization goals by anticipating in advance all the forms of
interactions that have to be supported. 

Web sites are not traditionally designed for
personalization. Enabling
true personalization implies supporting a flexible mode of interaction
between the user and the underlying information system. I hence adopt
the view that personalization is properly approached by studying interactions
of users with information systems~\cite{marchionni-book}. 

\section{Personalized Interaction}
There is a way by which a site can support both User 1 and User 2, 
without enumerating the scenario possibilities. The trick lies in recognizing
that both users bring different {\it partial information} to bear upon
the information-seeking activity. User 1 knows about manufacturer whereas
User 2 knows about lens type. Both attributes are only
partial specifications toward the selection of a camera and hence
a web site that flexibly allows the specification of partial information
will be personable to both users. Several techniques for exploiting
partial information are available. Two are described in this paper
--- partial evaluation (PE; a technique from the programming languages
community~\cite{jones}) and explanation-based generalization (EBG; a 
technique from the AI community~\cite{dejong}). Under certain 
situations~\cite{EBG_PE}, EBG and PE can be shown to
be essentially the same technique.

\subsection{Using Partial Evaluation}
\begin{figure}
\centering
\begin{tabular}{|l|l|} \hline
{\tt int pow(int base, int exponent) \{} & {\tt int pow2(int base) \{} \\
\,\,\,\,\,{\tt int prod = 1;} & \,\,\,\,\,{\tt return (base * base);} \\
\,\,\,\,\,{\tt for (int i=0;i<exponent;i++)} &  \} \\
\,\,\,\,\,\,\,\,\,\,{\tt prod = prod * base;} & \\
\,\,\,\,\,{\tt return (prod);} & \\
\} & \\
\hline
\end{tabular}
\caption{Illustration of the partial evaluation technique.
A general purpose {\tt pow}er function written in C (left) and
its specialized version (with {\tt exponent} statically set to 2) to handle
squares (right). Such specializations are performed automatically by partial
evaluators such as C-Mix.}
\label{pe}
\end{figure}

\begin{figure}
\centering
\begin{tabular}{|l|l|} \hline
{\tt if (Canon)} & \\
\,\,\,\,{\tt if (35mm)} & \\
\,\,\,\,{$\cdots \cdots \cdots$} & \\
\,\,\,\,{\tt else if (APS)} & \\
\,\,\,\,{$\cdots \cdots \cdots$} & \\
{\tt else if (Nikon)} & \\
\,\,\,\,{\tt if (35mm)} & \\
\,\,\,\,{$\cdots \cdots \cdots$} & {\tt if (Nikon)}\\
\,\,\,\,{\tt else if (APS)} & \,\,\,\,{$\cdots \cdots \cdots$} \\
\,\,\,\,{$\cdots \cdots \cdots$} & {\tt else if (Minolta)}\\
\,\,\,\,{\tt else if (SLR)} & \,\,\,\,{$\cdots \cdots \cdots$} \\
\,\,\,\,{$\cdots \cdots \cdots$} & \\
{\tt else if (Minolta)} & \\
\,\,\,\,{\tt if (35mm)} & \\
\,\,\,\,{$\cdots \cdots \cdots$} & \\
\,\,\,\,{\tt else if (SLR)} & \\
\,\,\,\,{$\cdots \cdots \cdots$} & \\
\hline
\end{tabular}
\caption{Using partial evaluation for personalization. (left) Programmatic 
input to partial evaluator.
(right) Specialized program from the partial evaluator, used to create 
a personalized information space.}
\label{cameras2}
\end{figure}

Partial evaluation is a technique for specializing computer programs, given
some (but not all) of their input. The input to a partial evaluator is
a program and some static information
about its arguments. The output of
the partial evaluator is a specialized version of this program (typically 
in the same language) where the static information has been used
to simplify
as many operations as possible. 
Partial evaluators analyze the given program to determine
reducible expressions and use
compiler transformations to interpret (evaluate) all such expressions.
Given the value of {\tt exponent} in
Fig.~\ref{pe} (left), the loop can be unrolled and the value of
the {\tt prod} variable forward propagated to yield 
the program in Fig.~\ref{pe} (right). Automatic partial evaluators are 
available for C, Scheme, PROLOG, and many other languages.

To use partial evaluation for personalization~\cite{naren-ic}, think of
a program as representing a dialog script of interaction between
a human and an information system. Every program variable is an opportunity for
the user to communicate a specification aspect.
For instance, we can abstract interaction with the browsing hierarchy in 
Fig.~\ref{cameras} (left) using the program of Fig.~\ref{cameras2} (left).
To personalize the site for User 2, we partially evaluate w.r.t. the
user's preferences (we set the program variable {\tt SLR} to 1 and all
conflicting variables such as {\tt 35mm} and {\tt APS} to zero). This
simplification, coupled with pruning dead-end branches (like the conditional
involving {\tt Canon}) leads to the specialized program of 
Fig.~\ref{cameras2} (right). A personalized web site can be recreated
from the simplified program. The act of partial evaluation thus mimics 
the processing of an out-of-turn input by the user.

\begin{figure}
\centering
\begin{tabular}{cc}
\includegraphics[width=2.4in]{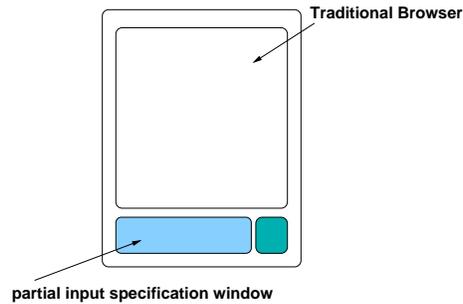}
\end{tabular}
\caption{Sketch of an interface for personalized interaction with web
sites.
The top window of the interface supports a traditional browsing
functionality. 
At any point in the
interaction, in addition, the user has the option of supplying personalization aspects
out-of-turn (bottom two windows). Such an interface can be implemented
as a toolbar in browsers.}
\label{toolbar}
\end{figure}
\begin{figure}
\centering
\begin{tabular}{cc}
\includegraphics[width=3in]{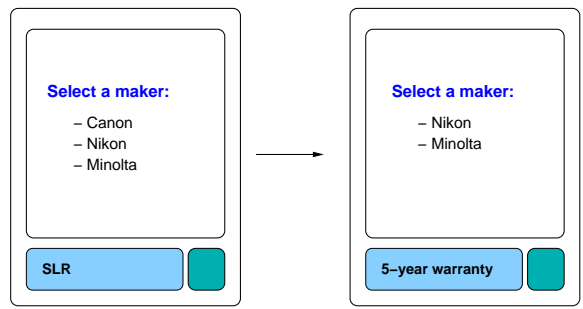}
\end{tabular}
\caption{Example of a personalized interaction for User 2. At the beginning of the
interaction, User 2 decides to not use any of the presented hyperlinks for camera model.
Instead, she uses the toolbar to specify her choice of camera type 
out-of-turn (left). The results of partial evaluation
cause the Canon option to be removed from the model choices (right).
Once again, User 2 opts to utilize the toolbar to provide a specification of warranty
information (results not shown).}
\label{sketchy2}
\end{figure}

This simple example shows that what is important is a
representation of interaction and an expressive operator (PE)
for supporting personalization. Our earlier statement that 
Fig.~\ref{cameras} (left) is unpersonable for User 2's activity implicitly
assumed that the operator was interpretation, i.e., values for program
variables are ascertained from the user in the order in which they were
modeled in the program. In reality, the notion of
representation of interaction was not a very useful one there.
With PE, both sites in Fig.~\ref{cameras} can be
made personable for both users. The notion of representation is crucial
here, since it drives the partial evaluation.

In human-computer interaction terms, the ability of the user to provide
an out-of-turn (but expected)
input is recognized as an important facet of {\it
mixed-initiative interaction}~\cite{allen-aimag,miimc}. In this view, we can think
of the act of clicking on a hyperlink as responding to the initiative
taken by the web site (information system). This is a system-directed
mode of interaction. Specifying a variable out-of-turn implies that the 
user has taken the initiative and made an unsolicited reporting.  This 
is a user-directed mode of interaction. Both these modes of interaction
are subsumed by PE, since they only differ in the arrival
time of specification aspects. We can thus perform a {\it sequence}
of partial evaluations to support a flexible interaction strategy
(see Figs.~\ref{toolbar} and~\ref{sketchy2}).
 
\subsection{Representations for Personalization}
We now formalize a definition to recognize the importance of
a representation:
\begin{descit}{(1)}
\noindent
A representation of interaction $\mathcal{I}$ is personable for an 
information seeking activity $\mathcal{G}$ if \\ $\mathcal{G}$ can be 
realized by a sequence of partial evaluations of $\mathcal{I}$.
\end{descit}

\noindent
We have shown that the representation of Fig.~\ref{cameras2} (left)
is personable
for both User 1 and User 2. Let us consider the usefulness of this
representation for some additional users:
\vspace{-0.3in}
\begin{descit}{}
\begin{description}
\item [User 3:] I am interested in cameras that allow the setting of shutter speed.
\item [User 4:] What types of cameras are available?
\item [User 5:] I am looking for a 35mm Canon.
\end{description}
\end{descit}

\noindent
As it is modeled in Fig.~\ref{cameras2} there are no program variables that
allow User 3 to describe his preference for shutter speed control. We say 
that the representation is unpersonable for this user's activity. This should 
be obvious, since personalization is captured by the assignment of values to
program variables. To support all personalization scenarios, we need to ensure
that adequate program variables are present in the representation.

Interestingly, the representation of Fig.~\ref{cameras2}
is unpersonable for Users 4 and 5 as well! User 4's request 
does involve camera types but this user is not expressing a selection 
of a type. Rather, he is initiating a dialog to discuss the specification of
camera type (which might eventually lead to a selection of type). 
We can use PE to specify a camera type out-of-turn but
we cannot use it to enquire about the types available. In general,
this requires the ability to reflect about the operation
of a program. 

One possibility is to explore other transformation techniques (e.g., slicing) that allow
the desired restructuring of the representation. Another 
is to retain PE as our personalization operator and change
the representation to accommodate User 4's request. This is a tricky endeavor
due to the optional nature of the request. 
The representation that
achieves this interaction is beyond the scope of this article; it involves
using continuations (e.g., via {\bf call/cc} in Scheme) to 
model the dynamic branching of control, in this case, to discuss the types of 
cameras available, and subsequently to return from such a dialog. For the
reader well versed in programming language implementation, this is akin to designing
exception handling capabilities (see Chapter 7 of~\cite{plbook}).

Let us now consider User 5's activity. This activity
specifies values for all applicable program variables, corresponding
to a {\it complete evaluation}, rather than a partial evaluation. Complete
evaluation implies that every aspect of interaction is determined in
advance, removing the need for any interaction! While we can accommodate
User 5's request, the representation of Fig.~\ref{cameras2} offers
no particular advantages for this activity (for instance, the
user could have just used the traditional, interpretive-style, browsing
interface). 

Thus, a representation can be unpersonable because there are
{\it no} program variables available to model user requests. Alternatively,
a representation is unpersonable because user requests can be captured 
only by specifying values for {\it all} variables. The crucial notion of
`values for {\it some, but not all} variables' is important to model 
personalized interaction by partial evaluation. This leads us to the next
two definitions:
\begin{descit}{(2)}
\noindent
A representation of interaction $\mathcal{I}$ is unpersonable for an 
information seeking activity $\mathcal{G}$ if \\ $\mathcal{G}$ cannot be 
realized by a sequence of partial evaluations of $\mathcal{I}$.
\end{descit}

\begin{descit}{(3)}
\noindent
A representation of interaction $\mathcal{I}$ is unpersonable for an 
information seeking activity $\mathcal{G}$ if \\ $\mathcal{G}$ can only be 
realized as a complete evaluation of $\mathcal{I}$.
\end{descit}
\noindent
Notice that definition {\it (3)} should be taken as the exception to our
earlier definition {\it (1)}.
\subsection{More on Representations}
We have thus far fixed the representation and asked the question:
`What are the activities for which a representation of interaction 
$\mathcal{I}$ is personable?'
This is the designer's viewpoint. For a given web site, it allows the designer 
to determine the target audience that will consider the site personable.

The complementary question is user-driven and asks:
`For my information seeking activity $\mathcal{G}$, what are the
personable representations?'
This allows the user to study different web site representations 
and see which ones lead to realization of
his information seeking goals.
In a realistic deployment of personalization solutions, many user studies
will likely be conducted while developing and designing the web site. A 
participatory
design methodology will attempt to address both perspectives.

\begin{figure}
\centering
\begin{tabular}{|ll|} \hline
click [here] & if you are interested in Nikon models.\\
click [here] & if you are interested in SLR cameras.\\
click [here] & if you are interested in cameras that allow shutter speed control.\\
click [here] & if you are seeking information about camera types.\\
click [here] & if you are interested in 35mm Canon models.\\
click [here] & $\cdots \cdots \cdots$ \\
\hline
\end{tabular}
\caption{`Freezing' interactions for all users results in an unpersonable
representation.}
\label{overfact}
\end{figure}

For instance, we have seen that the representation in 
Fig.~\ref{cameras2} (left) is
personable for Users 1 and 2 but unpersonable for Users 3, 4, and 5. 
Can we have a representation that is personable for all these users?
The trivial solution, which is to `freeze' and support all these situations,
actually leads to an unpersonable design! Fig.~\ref{overfact}
shows a design that allows all of these users to achieve their 
information seeking goals. However, there is really only one 
level of interaction, involving the choice of link to take.
All evaluations of the programmatic representation
have to be complete evaluations!
It can be argued that such an anticipatory design is unsatisfactory
since it buckets all users into predefined categories; it is 
hence not really `personal' at all! 

In practice we try to factor in different users' requirements into 
our modeling of interaction, without explicit enumeration. 
If the scenarios
requiring complete evaluation are a small fraction of the total number
of scenarios, then we can `excuse' our modeling 
for being unpersonable
for those situations (e.g., User 5). User 3's request can be accommodated by adding additional
program variables for shutter speed control. User
4's request cannot be easily accommodated, as stated earlier, without
significant revision of our programming style. If such requests will
be common, then we might be justified in redesigning the web site to
support such interactions. 

%\begin{table}
%\centering
%\begin{tabular}{|ll|} \hline
%{\bf Program modeling} & depth-first crawling; within-page modeling \\
%{\bf Variable assignment} & taxonomical relationships; domain semantics \\
%{\bf Web page reconstruction} & pruning dead-end branches \\
%\hline
%\end{tabular}
%\caption{Implementation decisions for using partial evaluation to
%personalize the Project VoteSmart web site.}
%\label{features-sammy}
%\end{table}

%\begin{figure}
%\centering
%\begin{tabular}{|c|c|} \hline
%maker & type \\
%\hline
%\end{tabular}
%\hspace{1in}
%\begin{tabular}{|c|c|c|} \hline
%state & party & branch \\ \hline
%      &       & seat \\
%\hline
%\end{tabular}
%\caption{Schemas of interaction at two web sites. (left) Fictitious
%camera retailer, (right) Project VoteSmart web site.}
%\label{lattice}
%\end{figure}
%

\subsection{Implementation Details}
The reader might notice that conditionals in Fig.~\ref{cameras2} such as
\begin{descit}{}
{\tt if (Canon)}
\end{descit}
\noindent
could be replaced by
\begin{descit}{}
{\tt if (manufacturer == Canon)}
\end{descit}

\noindent
Is one representation better than the other? In the original design, just 
setting `Canon' to 1 can model a user's request, but to be 
semantically correct we had to ensure that conflicting variables
such as `Minolta' and `Nikon' were set to zero (the {\tt else} clause will
trap some of these situations). This introduces some overhead in 
implementing a personalization system. In the new representation, we can merely
set the {\tt manufacturer} variable to `Canon' but the responsibility is on us
to realize that the user was referring to a camera manufacturer. 
For applications that have polysemous user input, this can be troublesome. 

The fundamental problem is learning a mapping from user requests to
assignments of values for internal representations (in our case,
program variables). This has been long studied in information retrieval 
in various forms, e.g., query expansion, local context analysis,
proximity metrics. In the personalization context, the assignments are
also expected to capture the semantics of out-of-turn interactions.
There are no solutions that work in general, only 
guidelines and recommendations.

As a demonstrator of the above ideas, we designed a personalization system
for the US Congressional portion of the Project VoteSmart web 
site (vote-smart.org). The site caters to information about
political individuals and users interact with the site
by specifying choices
of state, party, branch of congress, and seat. 
There are interesting dependencies 
underlying these attributes that cannot be captured by a 
clever representation. For instance, if the user says 
`Senior seat,' he is referring to a Senator, not a Representative.
Saying `North Dakota' and `Representative' in the current
political landscape defines a unique member of Congress (no party 
information is needed), and so on. It is important to consider such 
facets in order to deliver a compelling personalized experience. The net
effect of such considerations will be that multiple program variables are
initialized based on the user's input.

The unspecialized program was represented in an XML notation and modeled 
by a depth-first crawl of the site. 
%Thus every web
%page was a node (an if-statement) and the text anchoring the hyperlinks
%were used as the labels on edges (conditional expressions). 
Since
VoteSmart coalesces the party, branch, and seat information into a single
hyperlink, we conducted a within-page modeling to yield independently 
addressable program variables for these attributes. The partial 
evaluator was implemented using XSLT and a demo is available 
at pipe.cs.vt.edu. Evaluation results are presented in~\cite{pipe-tois}.

Additional web applications developed in this manner are described 
in~\cite{naren-ic,pipe-tochi}. Many of them conduct more sophisticated
modeling of interaction than just browsing hierarchies; for instance,
information integration from multiple web sites, interacting with
recommender systems, modeling clickable maps, and representing computed
information. All of these can be subsumed in a programmatic
modeling; to effect personalization, we just require
a meaningful way to view an interaction with an information system as a 
program and a mapping from user input to program variables. Opportunities to 
curtail the cost of partial evaluation 
for large information spaces are also studied in~\cite{pipe-tois}.  I refer 
the reader to these references for more discussion on implementation choices. 
In this article, I elaborate further on the usefulness of representations.

\section{More Choices of Representations}
Partial evaluation is one way to exploit partial information via a representation. 
Explanation-based generalization (EBG) is another. 
Even though they are computationally equivalent~\cite{EBG_PE}, we will
begin by making a distinction and later show the implications of their
equivalence for personalization.

With PE, a user experiences personalization because the site allows him to provide partial
information. With EBG, a user experiences personalization because the site knows some partial
information about him.
EBG is thus best understood here as a technique 
that incorporates 
partial information {\it prior} to a user interaction, whereas PE
incorporates partial information {\it during} a user's interaction.

%We introduce EBG by considering a very different form of personalization.
Consider a book-reader (Linus) revisiting the amazon.com website; a greeting prompts `Welcome 
back Linus.' After Linus selects a book for purchase, the website skips the
questions for credit card and shipping address when processing the order. This
is presumably because the answers to these parts of the interaction are being reused from a
previous session. 
%In particular, the site remembers that Linus likes to use his Discover
%card and prefers to ship books by Fedex. 
Admittedly, this is a useful form of personalization. 

%For variety, let us introduce another user (Sammy) who has never visited amazon.com. It is
%likely that the questions about credit card and shipping address cannot be skipped for Sammy. 

\begin{figure}
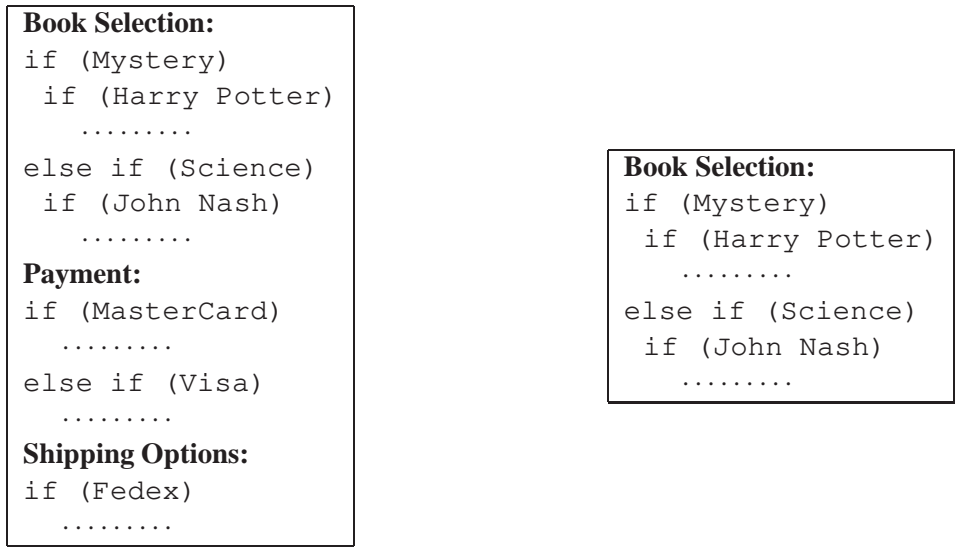

\centering
\begin{tabular}{lll}
\begin{tabular}{|l|} \hline
{\bf Book Selection:} \\
{\tt if (Mystery)} \\
\,\,\,\,{\tt if (Harry Potter)} \\
\,\,\,\,\,\,\,\,\,\,\,\,{$\cdots \cdots \cdots$} \\
{\tt else if (Science)} \\
\,\,\,\,{\tt if (John Nash)} \\
\,\,\,\,\,\,\,\,\,\,\,\,{$\cdots \cdots \cdots$} \\
{\bf Payment:} \\
{\tt if (MasterCard)} \\
\,\,\,\,\,\,\,\,{$\cdots \cdots \cdots$} \\
{\tt else if (Visa)} \\
\,\,\,\,\,\,\,\,{$\cdots \cdots \cdots$} \\
{\bf Shipping Options:} \\
{\tt if (Fedex)} \\
\,\,\,\,\,\,\,\,{$\cdots \cdots \cdots$} \\
\hline
\end{tabular}
& \hspace{1in} &
\begin{tabular}{|l|} \hline
{\bf Book Selection:} \\
{\tt if (Mystery)} \\
\,\,\,\,{\tt if (Harry Potter)} \\
\,\,\,\,\,\,\,\,\,\,\,\,{$\cdots \cdots \cdots$} \\
{\tt else if (Science)} \\
\,\,\,\,{\tt if (John Nash)} \\
\,\,\,\,\,\,\,\,\,\,\,\,{$\cdots \cdots \cdots$} \\
\hline
\end{tabular}
\end{tabular}
\caption{(left) Default interaction representation experienced by Amazon users.
(right) Interaction representation experienced by Linus. Lines such as `{\bf Payment:}'
are comments intended to show program structure.}
\label{linus-only}
\end{figure}

%Linus's interaction differs from Sammy's in that the partial information has been
%exploited even before he begins his current interaction with amazon.com. For Sammy, there is no such
%partial information that the site can exploit. From a representational viewpoint, the starting
%points for their interactions appear as shown in Fig.~\ref{linus-sammy}. Notice that from these
%respective starting points, either of Sammy and Linus can provide information out-of-turn (within the scope
%of the program variables available to them). 
%
%The key observation here is that Linus's starting representation can itself be viewed as a partial evaluation
%of Sammy's starting representation! Thus, partially evaluating the latter with respect to Linus's credit card
%and shipping address gives us Fig.~\ref{linus-sammy} (right). One can carry this analogy further and
%imagine a whole partial order of representations, beginning with the most general site, 
%defining various intermediate levels of representations, and continuing to the most specific,
%terminal, pieces of information. 

Fig.~\ref{linus-only} shows two representations, the default representation seen by Amazon users and the representation
experienced by Linus. It is as if the site has performed some `free' partial evaluations just for Linus!
According to our original definition, both representations are personable for Linus's
activity but Linus has to provide two extra pieces of information with the representation of Fig.~\ref{linus-only} (left).
Per EBG terminology, we say that there is a difference between them in terms of
{\it operationality}. Operationality deals with the issue of whether the site should remember 
Linus's credit card and payment information 
or whether it should require Linus to supply it during every interaction. This dilemma is actually 
at the heart of EBG.

\subsection{Using EBG}
Before we study EBG in more detail, we will make some preliminary observations. The above
dilemma is actually a dilemma for the designer of the personalization system and reduces to
the problem of identifying templates of interaction for users. 
A template --- such as the returning customer template ---
defines a starting point for a user interaction and identifies the program
variables that can be involved in the interaction. 
The tradeoff in designing templates is between the partial 
evaluations performed by the site (in the template) before the interaction begins and the partial evaluations 
conducted by the user during the interaction. 

%Both representations of Fig.~\ref{linus-only} 
%can be viewed as templates; of course, the left one is the vanilla template.
%Two such templates are shown in Fig.~\ref{linus-only}.

We can appreciate the difference by considering more users than just Linus.
If the design is set up so that the site performs most of the partial evaluations, then
a lot of templates will be needed to support all possible users. 
Each template provides a considerable amount of personalization 
but every user has to determine the right template for his
interactions. A mushrooming of template choices can cause frustrations
for users. Conversely, we can attempt to reduce the number of templates but
then some users might find that there is no template that {\it directly}
addresses their information-seeking goals. They might then proceed to use a default vanilla
template such as Fig.~\ref{linus-only} (left) (assuming that it is supported).
Such users may be able to satisfy their
goals but will experience longer interaction sequences and a not-so-personalized
interaction. The trick is to compress many intended scenarios
of interaction into a few template structures.

EBG is a systematic way to cluster the space of users and to determine dense regions
of repetitive interactions that could be supported. In Amazon, one important
distinction is that made between returning
customers and new customers. 
%In other words, there is a template for returning customers (such as Linus) and 
%a different template for other customers. 
The top-level prompt at the site makes this
distinction (this is automated with cookies) and transfers are made
to different interaction sequences.

How and why did Amazon decide on these two templates? Why not a distinction such as
`reading for pleasure versus reading for business or education?' Or, `students versus
professionals?' Two issues are important here.
First, given a customer, can the right template be determined {\it easily}? Determining if
a customer is a new or returning customer is admittedly easier to automate 
than determining if the person reads for pleasure! Second, the distinctions used for templating
interactions should translate into significantly different interaction sequences. Else, the
distinction is useless in practice. In the case of the returning customer, for instance,
Amazon can provide more personalized recommendations and exhibit a greater understanding of the 
customer's preferences and habits. Balancing these considerations is a long-studied problem in 
EBG; it is interesting that it surfaces in such a natural way in the personalization context.

At this point it should be clear that 
PE and EBG support different types of personalization.
While PE addresses the
expressiveness with which a user can supply partial information to the system, 
EBG addresses the expressiveness by which the system exploits 
partial information about the user. 

\begin{figure}
\centering
\begin{tabular}{cc}
\includegraphics[width=3in]{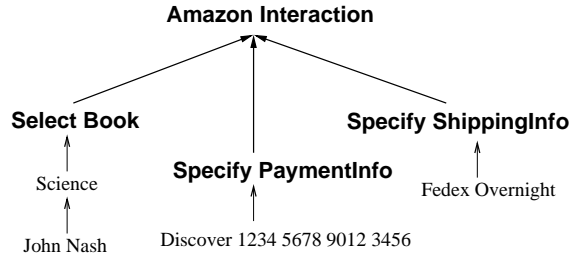}
\end{tabular}
\caption{Explaining a user's interaction as completing an information-seeking task.}
\label{op-slice}
\end{figure}

\begin{figure}
\centering
\begin{tabular}{cc}
\includegraphics[width=4.5in]{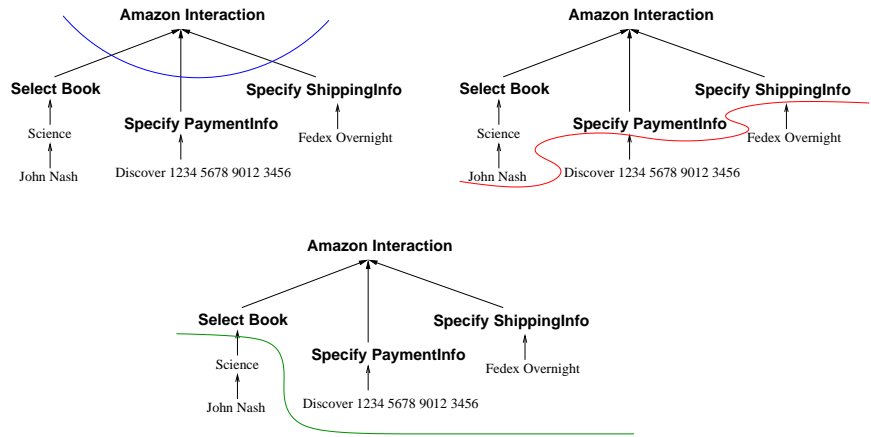}
\end{tabular}
\caption{Different choices of operationality boundaries lead to different 
templates of personalized interaction.}
\label{op-noslice}
\end{figure}

\subsection{Operationality Considerations}
EBG is an approach to reason from specific scenarios of interaction to general templates of interaction that should
be supported. A user's unpersonalized interaction with a web site is observed and a general
template is derived from it.
The first step is to use a domain theory to explain the user's interaction.
For our purposes, a domain theory captures the site layout, task models, browsing semantics, and
their role in information-seeking interactions. 
Explaining a user's successful interaction at a site with respect to the 
domain theory will help identify the parts of the interaction that contribute to achieving 
the personalization objectives. DeJong~\cite{dejong} shows
that an explanation can be viewed as a tree where each leaf is a
property of the example being explained, each internal node models
an inference procedure applied to its children, and the root is the
final conclusion supported by the explanation (namely, that the scenario was
an example of successful interaction). The explanation tree is used to
define a space of personable representations. Searching within this space is the
second step in EBG and is called operationalization.

Consider that Linus first used the Amazon site to select a book about John Nash (which he found by
browsing through the Science section of the site), paid with his Discover card, and chose Fedex
to ship the book. Explaining this interaction of Linus would lead to the proof tree shown in
Fig.~\ref{op-slice}. The tree shows how Linus satisfied the requirements of an
Amazon interaction; in this case, by satisfying the requirements for selecting a book,
specifying a payment information, and specifying his shipping details.
Each of these sub-requirements were in turn
satisfied by particular interaction sequences.
Operationalization can then be thought of as drawing a cutting plane
through the explanation tree. 
Every node below the plane is too specific
to be assumed to be part of all scenarios. The structure above the plane
is considered the persistent feature of all usage scenarios and is
expressed in the personalization system design. The user is then
expected to supply the details of the structure below the plane so that
the proof can be completed. Recall that since the proof below the plane is provided
by the user's clicks and selections, it can be performed in
a mixed-initiative manner.

Fig.~\ref{op-noslice} shows three ways of drawing a plane through the tree of Fig.~\ref{op-slice}.
The top left really draws the plane at the level of an Amazon interaction, implying that the site
will capture no personalization aspects. Every detail is meant to be supplied by the user in his
interaction. It is not even assumed, for instance, that the user will buy a book. This gives us
the vanilla template that caters to all users.
The top right of Fig.~\ref{op-noslice} draws the cutting plane to include the selection
of the book as subsumed by the system, leaving the payment and shipping address to be supplied by the
user. This is obviously a very strange notion of operationality! 
The template resulting from this option
would be appropriate only if
the same John Nash book is to be purchased over and over again with different credit card and shipment options!  
The bottom slice of Fig.~\ref{op-noslice} is probably the reasonable one where the payment and shipping
options are subsumed by the system, leaving the user to select the book. It 
recognizes the fact
that in a future interaction, the user is likely to purchase a different book.

Deriving a generalized template of interaction also depends on the class of users
it is intended to support. Is the template obtained from Linus supposed to apply only
to {\it his} future interactions or can it be applied to other users as well? 
Once again, there is a tradeoff. For instance, if we have multiple users in mind
then Fig.~\ref{op-noslice} (top right) no longer looks silly. Implementing this
template amounts to creating a `If you would like to buy the John Nash book, click here to
give payment options' link. Contrarily, Fig.~\ref{op-noslice} (bottom) would be strange here
since payment information and shipping details are not transportable across users.

After a template is derived, we have the option of explaining another user's 
interaction and deriving a new template, if this user's interaction is not
well captured by the existing template. 
As mentioned earlier, we need to be careful about an explosion in the 
number of templates if this process is repeated. 
Typically, the default vanilla representation is always retained
as one of the templates since there will be many users about whom the 
site has no prior information.

\subsection{Domain Theories for Information-Seeking Interactions} 
Operationality is thus a matter of utility and an example corresponds to a scenario 
of interaction. We can evaluate operationality choices by conducting usability studies and
determining the coverage of templates; example scenarios of interaction can be obtained by
observation and think-aloud protocols. But where do domain theories come from?

While there is significant understanding of information-seeking interactions, there are no
large, pertinent, domain theories available for the studies considered here. In~\cite{pipe-tochi},
we handcrafted a domain theory for reasoning about interactions at the `Pigments through the
Ages' website (http://webexhibits.org/pigments) and used it with EBG to design a personalization system.
At the end of this process, there is some optimism that domain theories can be prototyped for
certain recurring themes of information-seeking interactions. Besides supporting the construction of
explanations, domain theories can help in organizing software codebases for information system design.
In other application domains e.g., voice interface design and directory access protocols, this form of
codebase organization is already taking place. For instance, commercial speech recognition APIs provide
support for task-oriented dialogs (e.g., confirmations, purchase order processing) that make it easy
to prototype applications. Such an organization will greatly benefit the study of information
personalization.

%Note that EBG can be viewed as both {\it theory-driven generalization of examples} and
%{\it example-guided specialization of theories}~\cite{mitchell-book}. The former underscores the
%role of prior knowledge in inductive learning, namely how a domain theory will help us
%generalize from a user's interaction. The latter viewpoint corresponds more closely with
%personalization; it suggests that a user's example interaction is being used to create a personalization
%system (for future interactions).

\section{Personable Traits}
I have presented two views of personalization; both represent an information-seeking interaction 
and exploit partial information to deliver a customized experience. Together, they can help 
capture a variety of personalization scenarios. The EBG viewpoint is more prevalent than the
PE viewpoint because the way EBG harnesses partial information lends better to
implementation technologies. These observations point us to identifying the expressiveness in which partial 
information can be utilized by and communicated to an information system. 

\begin{figure}
\centering
\begin{tabular}{cc}
\includegraphics[width=3in]{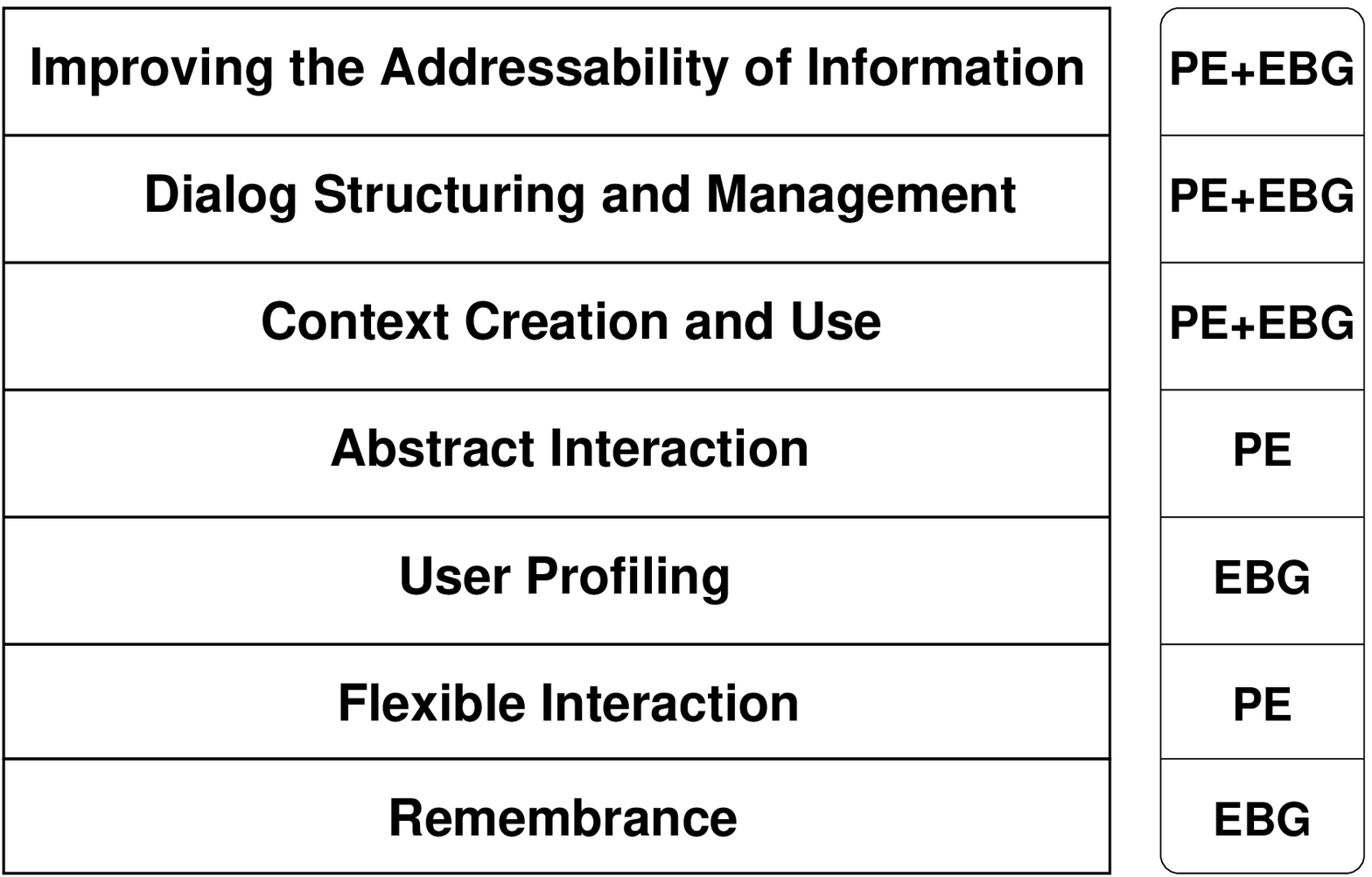}
\end{tabular}
\caption{Seven tiers of personalization, from simplest (bottom) to most sophisticated (top).}
\label{op-personal}
\end{figure}

In Fig.~\ref{op-personal},
I identify seven tiers of personable traits along such an axis, from most simple to most sophisticated. Alongside
each tier is also listed the primary way in which partial information is modeled and harnessed (PE or EBG or both).
In reading the following paragraphs, the reader should keep in mind that the presence of EBG is a situation
where the site knows something about the user whereas PE captures a situation where the user conveys something
to the site. It should also be remarked that many of the personalization solutions surveyed here do not
have explicit EBG or PE leanings; it is only our modeling of interaction that permits thinking of them in this
manner.

\subsection*{Remembrance}
This is an EBG mode of exploiting partial information and refers to the 
case where simple attributes of a user are remembered, such as credit cards and shipment 
options. Amazon is a prime example; Citibank Inc. used to provide a toolbar that provided
the same functionality. The partial information is thus being exploited in a per-user manner.
Web sites that capture and summarize simple form of interaction
history (e.g., top 10 visited pages) also 
fall into this category. Here, explanations from multiple user sessions are operationalized 
at the leaf level into a single template. This enables a type of personalization that is 
not specific to any user. For an EBG technique that can support this form of specialization,
see~\cite{flann-dietterich}.

\subsection*{Flexible Interaction}
This is a PE mode of personalization and supports simple forms of mixed-initiative interaction.
The partial information is expected to be supplied by the user and personalization enhances
the way in which it can be supplied. A good example is websites that allow the provision of
expected, but out-of-turn information, such as in the camera application described earlier.
Voice-activated systems are more advanced than websites
in their support for this type of personalization~\cite{miimc}.

\subsection*{User Profiling}
Our third tier is another example of EBG and is considerably more involved than remembrance.
Here, what the site knows about a user is not restricted to simple attribute-value information
but is actually a sophisticated model of prior interactions. For instance,
Amazon suggests `Since you liked Sense and Sensibility, you will also like Pride and Prejudice.' A user's
prior interaction is captured and explained. The explanation is operationalized at the level
of an internal representation, to be used in a future
interaction. This form of personalization has become very popular and many machine learning
techniques have been used to induce the internal representation (e.g., to learn a profile of
the user). Some of these techniques are now very sophisticated and try to work with many implicit
indicators. 

\subsection*{Abstract Interaction}
Just as user profiling extends remembrance in an EBG mode, abstract interaction extends flexible
interaction in a PE mode. Here the partial information that a user can supply is not restricted
to values for program variables but can be some abstract property of her interaction. For instance,
the user could be interested in movies that featured the lead actor in {\it Titanic}, but
may be unable to frame her partial information as `movies where Leonardo Di Caprio acted.' I am not
aware of any websites that provide such a functionality in any general way. Transformation
techniques for supporting such abstract interpretation are also scarce
(but see~\cite{ppe,jones-book}).

\subsection*{Context Creation and Use}
This tier of personalization involves both EBG and PE. An example is the shopping basket at Amazon
that allows a user to begin an interaction (PE) and save the state of 
the interaction to be
resumed later (EBG). When the user returns to the site, the shopping basket can be checked out
by providing the payment and delivery information. The ultimate goal of this tier is
to use context creation capabilities to help stage interactions. In many cases such staging naturally
breaks down into a context creation phase and a context usage phase.

\subsection*{Dialog Structuring and Management}
I have said that EBG and PE utilize partial information in different ways. However, if the
operationality boundary is moved down, then information meant to be supplied by the user becomes 
prior knowledge already known to the site. This shows that `designing a personalization system' versus `using a
personalizaton system' is quite an artificial distinction. The former just corresponds to
choosing a level of operationality (a partial evaluation, of the
domain theory), and the latter corresponds to
capturing user requests (again, via further partial evaluations, in this case of the template). This argument leads to the equivalence between
EBG and PE established in~\cite{EBG_PE}. This tier of personalization removes
the distinction between EBG and PE and the interaction resembles more a dialog,
with all the associated benefits of a conversational mode. There are not many websites that
support such a tier of personalization but this problem has been studied in other delivery
mechanisms such as speech technologies~\cite{allen-aimag}.

\subsection*{Improving the Addressability of Information}
The holy grail of personalization is to provide constructs that improve the addressability of
information. Consider how a person can communicate the homepage of {\it AI Magazine} to another.
One possibility is to specify the URL; in case the reader is unaware, the URL is quite lengthy.
Another is to just say ``Goto google.com, type {\it AI Magazine}, and click the `I'm feeling
Lucky' button.'' The advantage of the latter form of description is that it enhances the addressability 
of the magazine's webpage, by using terms already familiar to the visitor. This tier of personalization thus 
involves determining and reasoning about the addressability of information as a fundamental goal, 
before attempting to deliver personalization. All the previous tiers have made implicit assumptions
about addressability. Solutions in this tier take into account various criteria from the user (or learn it automatically from interactions)
and use them to define and track addressability constraints. Such information is
then used to support personalization. This helps exhibit a deeper understanding of how the user's
assumptions of interaction dovetail with his information-seeking goals. The first steps toward
understanding addressability have been taken~\cite{dumais}. However, the modeling of interaction here
assumes a {\it complete information} view, rather than partial information.

\section{Discussion}
My view of personalization is admittedly a very simple one. It only aims to capture the
{\it interaction} aspects underlying a personalized experience and not many others such 
as quality, speed, and utility. For instance, Amazon's recommender might produce better
recommendations than some other bookseller's but if they have the same interaction sequences, then
the modeling methodology presented here cannot distinguish between them. 
The contribution of the methodology is that by 
focusing solely on modeling interaction, it provides 
a vocabulary for reasoning about information-seeking.  One
direction of future work is to prototype software tools to support
the types of analyses discussed above (in a manner akin to~\cite{footprints}).

While I have resisted the temptation to unify all meanings of the word `personalization,'
I will hasten to add that EBG and PE are only two ways of harnessing partial information.
Any other technique that addresses the capture, modeling, or processing of partial information
in the context of interactions will readily find use as the basis for a
personalization system. The operative keyword here is, thus, {\it partial}.
A long-term goal is to develop a theory of reasoning about representations of information systems. The
ideas presented here provide a glimpse into what such a theory might look like.

\section*{Acknowledgements}
This article is a gift to my mother, who suggested the title.
My view of personalization has benefited from interactions with
Rob Capra, Sammy Perugini, Manuel Perez, and Priya Kandhadai.

\bibliographystyle{plain}

\end{document}